\documentclass{article} 
\usepackage[preprint]{colm2026_conference}

\usepackage{microtype}
\usepackage{hyperref}
\usepackage{url}
\usepackage{booktabs}
\usepackage{wrapfig}
\usepackage{amsmath}
\usepackage{amssymb}
\usepackage{bm}
\usepackage{amsfonts}
\usepackage{mathtools}
\usepackage{algorithm}
\usepackage[noend]{algpseudocode}
\usepackage{caption}
\usepackage{multirow}
\usepackage{xcolor}
\usepackage{graphicx}
\usepackage{bbding}
\algrenewcommand\algorithmiccomment[1]{\hfill$\triangleright$~#1}

\usepackage{lineno}

\definecolor{darkblue}{rgb}{0, 0, 0.5}
\hypersetup{colorlinks=true, citecolor=darkblue, linkcolor=darkblue, urlcolor=darkblue}
\newcommand{\AlgoTopRule}{\par\noindent\rule{\linewidth}{0.8pt}}
\newcommand{\AlgoMidRule}{\par\noindent\rule{\linewidth}{0.5pt}}
\newcommand{\AlgoBottomRule}{\par\noindent\rule{\linewidth}{0.8pt}}

\title{BitLM: Unlocking Multi-Token Language Generation with Bitwise Continuous Diffusion}

\author{
  \normalfont
  \begin{minipage}{0.98\textwidth}
  \centering
  {\normalsize
  Shaobin Zhuang\textsuperscript{1},
  Yuang Ai\textsuperscript{2,3},
  Jiaming Han\textsuperscript{2},
  Xiaohui Li\textsuperscript{1},
  Huaibo Huang\textsuperscript{3} \\
  Xiangyu Yue\textsuperscript{2},
  Xuefeng Hu,
  Kun Xu,
  Yali Wang\textsuperscript{4}\Envelope,
  Hao Chen\Envelope
  }\\[-0.0em]
  {\footnotesize
  \textsuperscript{1}Shanghai Jiao Tong University \quad
  \textsuperscript{2}MMLab, The Chinese University of Hong Kong \quad \\
  \textsuperscript{3}Institute of Automation, Chinese Academy of Sciences \quad
  \textsuperscript{4}SIAT, Chinese Academy of Sciences
  }
  \end{minipage}
  }

\begin{document}

\ifcolmsubmission
\linenumbers
\fi

\maketitle

\begin{abstract}

Autoregressive language models generate text one token at a time, yet natural language is inherently structured in multi-token units—phrases, n-grams, and collocations that carry meaning jointly. This one-token bottleneck limits both the expressiveness of the model during pre-training and its throughput at inference time. Existing remedies such as speculative decoding or diffusion-based language models either leave the underlying bottleneck intact or sacrifice the causal structure essential to language modeling. We propose \textbf{BitLM}, a language model that represents each token as a fixed-length binary code and employs a lightweight diffusion head to denoise multiple tokens in parallel within each block. Crucially, BitLM preserves left-to-right causal attention across blocks while making joint lexical decisions within each block, combining the reliability of autoregressive modeling with the parallelism of iterative refinement. By replacing the large-vocabulary softmax with bitwise denoising, BitLM reframes token generation as iterative commitment in a compact binary space, enabling more efficient pre-training and substantially faster inference—without altering the causal foundation that makes language models effective. Our results demonstrate that the one-token-at-a-time paradigm is not a fundamental requirement but an interface choice, and that changing it can yield a stronger and faster language model. We hope BitLM points toward a promising direction for next-generation language model architectures.

\end{abstract}
\section{Introduction}
\label{sec:intro}
Modern large language models are usually described as models of language, but operationally they are models of transitions between vocabulary IDs. Given a prefix, a transformer computes a contextual hidden state and converts it into a normalized distribution over a vocabulary; one token is then sampled, appended to the prefix, and the process repeats. This next-token paradigm has underwritten the success of contemporary LLMs across scales and domains \citep{brown2020language,chowdhery2023palm,touvron2023llama,grattafiori2024llama}. At the same time, it imposes a very specific interface between hidden-state computation and symbolic output: language must be emitted as a sequence of atomic categorical decisions. That interface has long been recognized as restrictive in its own right \citep{yang2017breaking}, and it also makes generation intrinsically sequential at inference time.

A large body of recent work has tried to reduce this sequential bottleneck without changing the underlying output interface. Blockwise decoding, speculative decoding, and multi-token prediction accelerate generation by proposing or verifying several future tokens at once \citep{stern2018blockwise,leviathan2023fast,cai2401medusa,gloeckle2024better}. 
Non-autoregressive and semi-autoregressive methods go further by relaxing strict left-to-right factorization \citep{gu2017non,wang2018semi,ghazvininejad2019mask}. In parallel, diffusion and iterative refinement have emerged as compelling alternatives to one-step categorical prediction for discrete generation \citep{austin2021structured,li2022diffusion,gong2022diffuseq,lou2024discrete}. These directions have produced important insights, but they often leave one assumption untouched: the model still ultimately talks to text through vocabulary-level categorical outputs, whether one token at a time or many.

\begin{wrapfigure}{r}{0.48\textwidth}
  \vspace{-1.5em} 
  \centering
  \includegraphics[width=\linewidth]{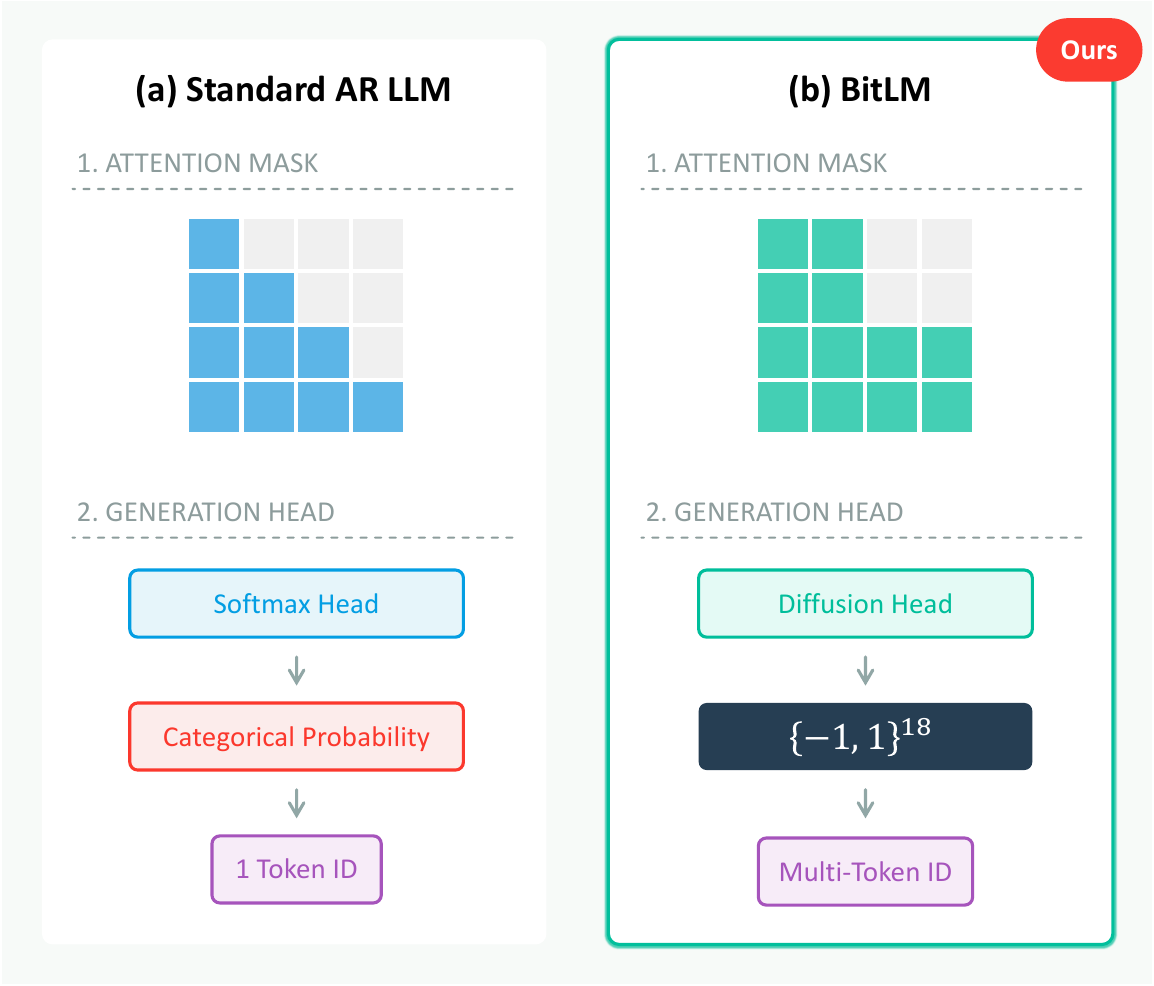}
  \caption{\textbf{Conceptual comparison between standard AR LLMs and BitLM}. By replacing the softmax head with a diffusion head, BitLM reformulates token generation as iterative denoising in a compact binary space, enabling multi-token joint realization.}
  \label{fig:teaser}
  \vspace{-1em} 
\end{wrapfigure}

This paper starts from a different premise, as conceptually illustrated in Figure~\ref{fig:teaser}. Instead of asking how to decode the vocabulary softmax more efficiently, we ask whether the vocabulary softmax should be the primary interface at all. A token identity is usually treated as an atomic class, but it can equally be represented as a short binary code. On the surface, this may look like a simple reparameterization. Our claim is that it changes the geometry of generation. A vocabulary softmax places lexical alternatives on a simplex and encourages per-position categorical decisions; a binary representation turns token prediction into denoising over a small set of coordinated binary variables. In that view, generation is no longer only the problem of selecting the next token ID. It becomes the problem of realizing a continuation as a structured discrete object that can be refined jointly across multiple positions.

This perspective resonates with several recent developments. Byte-level and token-free models have shown that the basic interface between raw text and sequence modeling is more flexible than wordpiece tokenization might suggest \citep{xue2022byt5,yu2023megabyte}. Diffusion-based models have shown that iterative refinement can be a powerful generative principle for discrete data \citep{austin2021structured,li2022diffusion,gong2022diffuseq,lou2024discrete}. In vision and other discrete domains, binary formulations such as Analog Bits demonstrate that diffusion in bit space can provide a simple and effective bridge between discrete symbols and continuous denoising dynamics \citep{chen2022analog}. Yet this perspective remains underexplored in language modeling, where the output layer is still overwhelmingly organized around a large categorical vocabulary distribution. We argue that language is precisely the setting where this question matters most, because the output interface is not only a modeling choice but also a decoding bottleneck.

Motivated by this observation, we propose \textbf{BitLM}, which transfers the binary-space denoising perspective of recent visual token generation methods such as BitDance \citep{ai2026bitdance, zhuang2026uniwetok} to text. As shown in Figure~\ref{fig:teaser}, BitLM represents each tokenizer entry as a fixed-length binary code, projects these codes into the hidden dimension of a standard LLM backbone, and performs causal computation in the usual left-to-right manner. Instead of decoding with a vocabulary softmax, the model uses the resulting contextual latent to condition a diffusion head that jointly denoises the binary codes of future tokens. Because the diffusion head operates over multiple positions simultaneously, it can realize a block of tokens in parallel rather than treating each future position as an isolated categorical draw. By adjusting the attention mask from fully causal to block-causal, the model can vary the degree of parallelism while preserving causal dependencies across blocks.

The resulting picture is conceptually simple: the backbone is responsible for \emph{reasoning about what should happen next}, while the diffusion head is responsible for \emph{realizing how that continuation should crystallize into discrete symbols}. This separation is the central idea of BitLM. It preserves the causal inductive bias that has made autoregressive LLMs so effective, but replaces the final vocabulary-level decision with joint denoising in binary space. Unlike post-hoc acceleration schemes, blockwise parallel generation is therefore not an external add-on; it is a native consequence of the model's generative interface. More broadly, BitLM suggests that the large-vocabulary softmax is a historical choice rather than a necessary endpoint of language modeling.

Our goal is not to claim that binary coding is a universal replacement for token-level language modeling. Rather, we use BitLM to expose a broader design dimension that has received relatively little attention: the geometry of the symbolic output space. Once that geometry is changed, new decoding regimes become natural. In particular, block-causal generation can be understood not as an approximation to autoregressive sampling, but as a first-class mode of generation in its own right. This viewpoint provides a clean conceptual bridge between three literatures that are often discussed separately: autoregressive language modeling, diffusion-based discrete generation, and parallel decoding.

This paper makes three main contributions. First, it introduces a binary-code formulation of language modeling that replaces vocabulary-level prediction with denoising in a bit space. Second, it proposes a simple architecture that combines a causal LLM backbone with a diffusion head for joint blockwise lexical realization, naturally enabling block-causal parallel generation. Third, it advances a broader perspective on language generation: changing the geometry of the output space can change the geometry of decoding itself.
\section{Related Work}
\label{sec:related}


\paragraph{Parallel and semi-autoregressive generation.}
A large body of work has tried to reduce the serial bottleneck of next-token decoding without changing the underlying vocabulary-level categorical interface. Foundational non-autoregressive, semi-autoregressive, and iterative-refinement models relax strict left-to-right factorization through parallel token prediction or repeated editing \citep{gu2017non,wang2018semi,ghazvininejad2019mask,stern2019insertion}. More recent LLM-oriented methods typically keep the standard language-model head and accelerate inference through better proposal or verification mechanisms, including blockwise parallel decoding \citep{stern2018blockwise}, speculative decoding \citep{leviathan2023fast}, auxiliary multi-token heads such as Medusa \citep{cai2401medusa}, and multi-token prediction objectives \citep{gloeckle2024better}. BitLM is aligned with this literature in its goal of exposing more parallelism at generation time, but it differs in where that parallelism enters the model. Instead of proposing several categorical tokens and then accepting, rejecting, or verifying them under a conventional softmax head, BitLM realizes an entire future block by jointly denoising binary token codes. Blockwise generation is therefore a property of the model's native output interface rather than an external decoding procedure. Conversely, unlike exact speculative methods, BitLM does not aim to preserve the distribution of a pre-existing autoregressive model; it defines a different generative parameterization.

\paragraph{Diffusion and iterative refinement for text.}
A second line of work models language generation as iterative denoising. Early work established general diffusion processes in discrete state spaces \citep{austin2021structured}. Subsequent text diffusion models denoised continuous embeddings \citep{li2022diffusion,strudel2022self,gong2022diffuseq}, introduced autoregressive or semi-autoregressive diffusion factorizations \citep{wu2023ar,han2023ssd}, and developed stronger discrete or score-based parameterizations \citep{zheng2023reparameterized,lou2024discrete,sahoo2024simple}. Recent large-scale variants such as LLaDA and Block Diffusion further narrowed the gap between diffusion and autoregressive language modeling while enabling flexible-length or blockwise generation \citep{nie2025large,arriola2025block}. BitLM is closest to the semi-autoregressive and block-diffusion part of this literature, but differs in the space in which denoising occurs. Prior text diffusion models typically refine embeddings, masks, simplex states, or categorical token variables; BitLM instead denoises fixed-length binary token codes and uses diffusion only for lexical realization on top of a causal LLM backbone. In this sense, the contextual computation remains a standard left-to-right transformer computation, while the final symbolic realization is moved into an iterative binary space. We do not view this as a universal replacement for vocabulary-space or mask-space diffusion. Rather, it is a different point in the design space that makes joint within-block lexical sampling particularly natural.

\paragraph{Alternative symbolic interfaces and output layers.}
A third line of work questions whether the large-vocabulary softmax should be treated as the only interface between hidden states and symbols. Theoretical and architectural studies have analyzed limitations of the conventional output layer \citep{yang2017breaking}. Earlier large-vocabulary methods such as binary code prediction replaced flat softmax with bitwise classifiers to reduce output-layer cost \citep{oda2017neural}. Orthogonally, token-free or byte-level models such as ByT5, MEGABYTE, and BLT alter the primitive modeling unit itself rather than the decoding rule \citep{xue2022byt5,yu2023megabyte,pagnoni2025byte}. Most directly relevant to our work are binary-space generative models. Analog Bits showed that discrete symbols can be represented as fixed-length binary codes and generated by continuous denoising \citep{chen2022analog}, while recent visual-token work such as BitDance and UniWeTok demonstrated the promise of very large binary code spaces for image and multimodal generation \citep{ai2026bitdance,zhuang2026uniwetok}. BitLM transfers this binary denoising perspective to language, but with two important differences. Unlike earlier binary output layers for text, it does not treat bits as independent one-shot classification targets. Unlike visual binary-token models, it couples binary-space generation with a causal language backbone and a block-causal generation regime. In this sense, BitLM is less an output-layer compression method than an exploration of a different symbolic geometry for language generation.

Overall, BitLM sits at the intersection of these three directions. It inherits causal context computation from autoregressive language models, joint iterative realization from diffusion models, and a binary symbolic interface from analog-bit generation, while using block-causal factorization to make parallel decoding native to the model rather than an after-the-fact approximation. We therefore view it as complementary to, rather than a drop-in replacement for, mature softmax-based autoregressive systems.
\begin{figure*}
    \centering
    \includegraphics[width=\linewidth]{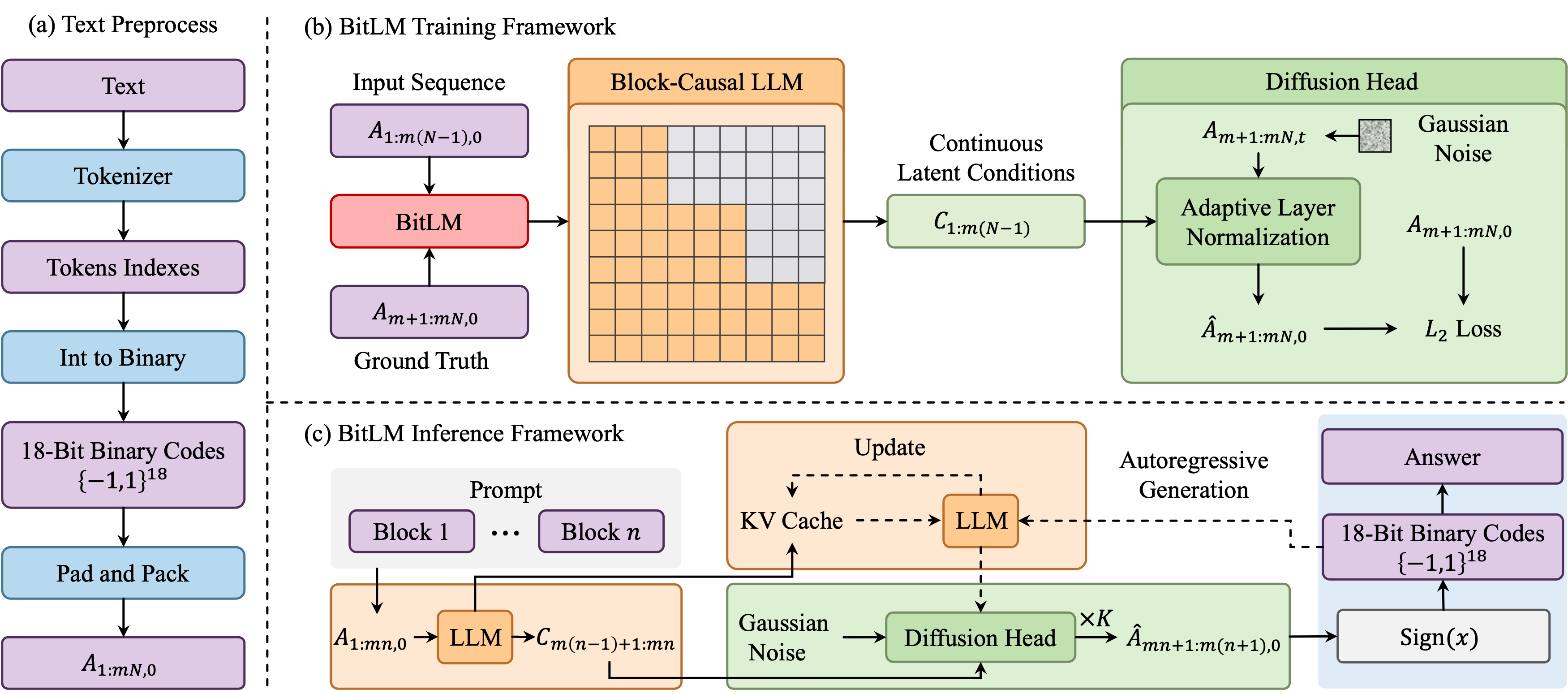}
    \caption{\textbf{Illustration of the training and inference frameworks of BitLM.}
    }
    \label{fig:framework}
\end{figure*}

\section{Method}
\label{sec:method}

BitLM replaces the conventional vocabulary softmax with conditional denoising in a fixed binary token space. The model keeps a causal language backbone for contextual computation, but realizes the next block of tokens through iterative denoising of binary codes. This yields a block-level factorization
\begin{equation}
p(y_{1:L}) = \prod_{n=1}^{N} p\!\left(y^{(n)} \mid y^{(<n)}\right),
\label{eq:block_factorization}
\end{equation}
where $y_{1:L}$ is a token sequence, $y^{(n)}$ denotes the $n$-th block of $m$ tokens, and $N=\lceil L/m\rceil$ is the number of blocks after padding. The key change is that each factor in Eq.~\eqref{eq:block_factorization} is modeled in binary space rather than by $m$ independent vocabulary softmaxes.

\subsection{Binary Token Interface}
\label{subsec:binary_interface}

Let $\mathcal{V}$ be a tokenizer vocabulary of size $V$, and let
\begin{equation}
B = \left\lceil \log_2 V \right\rceil
\end{equation}
be the binary code length. In our experiments $B=18$, so that $2^B$ covers the tokenizer index range. We keep the tokenizer itself unchanged and only replace the symbolic output interface. Concretely, each token id $y_i \in \{0,\dots,V-1\}$ is mapped to a fixed $B$-bit binary code,
\begin{equation}
\phi(y_i) = 2 \cdot \operatorname{bin}_B(y_i) - 1 \in \{-1,1\}^{B},
\label{eq:binary_code}
\end{equation}
where $\operatorname{bin}_B(\cdot)\in\{0,1\}^B$ returns the $B$-bit integer representation. The $0 \mapsto -1$ transformation places token symbols on the vertices of a binary hypercube, following the continuous binary-space view used in prior discrete diffusion work \citep{chen2022analog,ai2026bitdance}.

For a sequence $y_{1:L}$, we stack all token codes into
\begin{equation}
A_{1:L,0} = \big[\phi(y_1),\phi(y_2),\dots,\phi(y_L)\big]^\top \in \{-1,1\}^{L \times B},
\label{eq:clean_binary_sequence}
\end{equation}
where the subscript $0$ denotes clean binary codes.
We then pad the sequence to a multiple of the block size $m$, obtaining a padded tensor
\begin{equation}
A_{1:Nm,0} \in \{-1,1\}^{Nm \times B}, \qquad N=\left\lceil \frac{L}{m} \right\rceil,
\end{equation}
and partition it into $N$ blocks:
\begin{equation}
A^{(n)}_0 = A_{m(n-1)+1:mn,\,0} \in \{-1,1\}^{m \times B}, \qquad n=1,\dots,N.
\label{eq:block_binary_sequence}
\end{equation}
Only the first $V$ binary patterns are assigned to tokenizer ids; any remaining patterns in $\{-1,1\}^B$ are left unused. Importantly, the binary code here is a fixed identifier rather than a learned semantic codebook. This lets us isolate the effect of changing the \emph{output interface} while keeping the text tokenizer unchanged.

To feed binary tokens into the backbone, we lift the per-token channel dimension from $B$ to the hidden size $d$ using a position-wise MLP:
\begin{equation}
E_{1:L} = \operatorname{MLP}\!\left(A_{1:L,0}\right) \in \mathbb{R}^{L \times d}.
\label{eq:binary_lifting}
\end{equation}
This projection replaces the usual token embedding lookup and maps each binary token into the backbone hidden space.

\subsection{Conditional Denoising in Binary Space}
\label{subsec:binary_denoising}

BitLM models each target block as a denoising problem \citep{lipman2022flow,ho2020denoising,song2020score} in continuous binary space. Given a clean target block $A^{(n)}_0 \in \{-1,1\}^{m \times B}$, we sample a timestep $t \sim \mathcal{U}[0,1]$ and Gaussian noise
\begin{equation}
\epsilon \sim \mathcal{N}(\mathbf{0}, \mathbf{I}_{m \times B}),
\end{equation}
and construct a noisy analog-bits state by straight-line interpolation:
\begin{equation}
A^{(n)}_t = (1-t) A^{(n)}_0 + t \epsilon \in \mathbb{R}^{m \times B}.
\label{eq:forward_noising}
\end{equation}
Thus $t=0$ corresponds to a clean binary block and $t=1$ corresponds to pure Gaussian noise.
Given a contextual condition tensor $C^{(n-1)} \in \mathbb{R}^{m \times d}$ for the next block, the diffusion head predicts the clean block from its noisy version:
\begin{equation}
\hat{A}^{(n)}_0 = \operatorname{DiffHead}_\theta\!\left(A^{(n)}_t,\, t;\, C^{(n-1)}\right) \in \mathbb{R}^{m \times B}.
\label{eq:diff_head}
\end{equation}
We keep the denoiser itself deliberately lightweight; the contribution of BitLM lies in the binary generative interface and the block-causal factorization, rather than in a specialized diffusion architecture.

The contextual latent is injected into the denoiser through adaptive layer normalization (AdaLN). Let $h \in \mathbb{R}^{m \times d_h}$ be an intermediate hidden state in the diffusion head, and let $e(t)\in\mathbb{R}^{d}$ be a timestep embedding broadcast over the $m$ positions. We use token-wise conditioning of the form
\begin{equation}
\operatorname{AdaLN}(h; C^{(n-1)}, t)
=
\gamma\!\left(C^{(n-1)}, e(t)\right)\odot \operatorname{LN}(h)
+
\beta\!\left(C^{(n-1)}, e(t)\right),
\label{eq:adaln}
\end{equation}
where $\gamma(\cdot),\beta(\cdot)\in\mathbb{R}^{m \times d_h}$ are learned affine modulation functions and $\odot$ denotes element-wise multiplication.
Follow previous works \citep{ai2026bitdance,li2025back,yu2025pixeldit},
we train the denoiser with a simple $x_0$-prediction:
\begin{equation}
\mathcal{L}_{\mathrm{diff}}^{(n)}
=
\mathbb{E}_{t,\epsilon}
\left[
\left\|
\hat{A}^{(n)}_0 - A^{(n)}_0
\right\|_2^2
\right].
\label{eq:diffusion_loss}
\end{equation}
Following Analog Bits and BitDance, after iterative denoising we apply a hard sign projection to return to the discrete binary space \citep{chen2022analog,ai2026bitdance}:
\begin{equation}
\bar{A}^{(n)}_0 = \operatorname{sign}\!\left(\hat{A}^{(n)}_0\right) \in \{-1,1\}^{m \times B}.
\label{eq:sign_projection}
\end{equation}
The inverse map $\phi^{-1}$ simply interprets $-1/+1$ as $0/1$, reconstructs the integer token ids, and converts them back to text tokens.

\subsection{Block-Causal Context Computation}
\label{subsec:block_causal_backbone}

The language backbone computes contextual states over the lifted binary sequence using a block-causal attention mask. Let
\begin{equation}
C_{1:L} = \operatorname{BlockCausalLLM}\!\left(E_{1:L}; \mathcal{M}_m\right) \in \mathbb{R}^{L \times d},
\label{eq:block_causal_llm}
\end{equation}
where $\mathcal{M}_m$ is a block-causal mask with block size $m$. 
At inference time, we use the same notation for the incremental cached forward pass of the backbone. When a KV cache is present,
$\operatorname{BlockCausalLLM}(\cdot; \mathcal{M}_m, \mathrm{KV})$
denotes a single-block update that returns the new block hidden states together with the updated cache.
Define the block index of position $i$ as
\begin{equation}
b(i) = \left\lfloor \frac{i-1}{m} \right\rfloor + 1,
i\in\{1,2,\dots,L\}.
\end{equation}
Then the attention mask is
\begin{equation}
(\mathcal{M}_m)_{ij}
=
\begin{cases}
0, & b(j) \le b(i),\\
-\infty, & b(j) > b(i).
\end{cases}
\label{eq:block_causal_mask}
\end{equation}
Unlike a standard causal mask, Eq.~\eqref{eq:block_causal_mask} allows all tokens inside the same block to attend to one another, while preserving causal dependence across blocks. When $m=1$, this reduces exactly to standard left-to-right causal attention.

We use the hidden states of block $n\!-\!1$ to condition generation of block $n$. Let
\begin{equation}
C^{(n)} = C_{m(n-1)+1:mn} \in \mathbb{R}^{m \times d}
\end{equation}
be the contextual states of block $n$. Then the next-block distribution is parameterized as
\begin{equation}
p_\theta\!\left(y^{(n)} \mid y^{(<n)}\right)
\equiv
p_\theta\!\left(A^{(n)}_0 \mid C^{(n-1)}\right),
\label{eq:block_prediction}
\end{equation}
with $C^{(0)}$ provided by a learned BOS block or by the external prompt. Operationally, this means training targets are shifted by one block: the backbone summarizes the current realized block, and the diffusion head uses that summary to realize the next one. This one-block shift keeps training and inference aligned.

\subsection{Joint Block Realization}
\label{subsec:joint_block_realization}

A central property of BitLM is that the diffusion head predicts the entire target block jointly. The input to Eq.~\eqref{eq:diff_head} is the full noisy tensor $A^{(n)}_t \in \mathbb{R}^{m \times B}$, and the output is the full denoised tensor $\hat{A}^{(n)}_0 \in \mathbb{R}^{m \times B}$. Consequently, the model does \emph{not} impose a factorization of the form \citep{devlin2019bert,nie2025large, bie2025llada2}
\begin{equation}
p_\theta\!\left(A^{(n)}_0 \mid C^{(n-1)}\right)
=
\prod_{i=1}^{m}\prod_{b=1}^{B}
p_\theta\!\left(a^{(n)}_{i,b} \mid C^{(n-1)}\right),
\label{eq:no_factorization}
\end{equation}
which would correspond to independent sampling across positions and bit channels. Instead, inter-token and inter-bit dependencies inside the block are represented implicitly by the denoising dynamics. This is precisely why the binary diffusion head is a natural fit for block-causal parallel generation.

The full training objective averages Eq.~\eqref{eq:diffusion_loss} over all valid blocks:
\begin{equation}
\mathcal{L}_{\mathrm{BitDance\text{-}LM}}
=
\frac{1}{N}
\sum_{n=1}^{N}
\mathcal{L}_{\mathrm{diff}}^{(n)}.
\label{eq:full_loss}
\end{equation}
Algorithm~\ref{alg:bitdance_train} summarizes the end-to-end training pipeline of BitLM, from binary encoding and block-causal context computation to conditional denoising of the next block.



\begin{figure*}[t]
\centering

\begin{minipage}[t]{0.485\textwidth}

\captionof{algorithm}{BitDance-LM training.}
\label{alg:bitdance_train}

\AlgoTopRule

\small
\begin{algorithmic}[1]
\Require Token sequence $y_{1:L}$, block size $m$, code length $B$
\Ensure Training loss $\mathcal{L}_{\mathrm{BitDance\text{-}LM}}$

\AlgoMidRule

\State $N \gets \lceil L / m \rceil$
\State $A_{1:L,0} \gets \phi(y_{1:L})$ \Comment{token ids $\rightarrow$ binary codes in $\{-1,1\}^{B}$}
\State Pad $A_{1:L,0}$ to length $Nm$, yielding $A_{1:Nm,0}$
\State Split $A_{1:Nm,0}$ into blocks $\{A_0^{(1)}, \ldots, A_0^{(N)}\}$, where $A_0^{(n)} \in \{-1,1\}^{m \times B}$
\State $E_{1:Nm} \gets \operatorname{MLP}(A_{1:Nm,0})$ \Comment{lift binary codes to hidden space}
\State Compute blockwise context states $\{C^{(1)}, \ldots, C^{(N)}\}$ with
\Statex \hspace{\algorithmicindent} $\operatorname{BlockCausalLLM}(E_{1:Nm}; \mathcal{M}_m)$
\State Initialize $C^{(0)}$ from a learned BOS block or the external prompt
\State $\mathcal{L} \gets 0$

\For{$n = 1, \ldots, N$}
    \State $t \sim \mathcal{U}[0,1]$, \quad $\epsilon \sim \mathcal{N}(0, I_{m \times B})$
    \State $A_t^{(n)} \gets (1-t) A_0^{(n)} + t \epsilon$
    \State $\hat{A}_0^{(n)} \gets \operatorname{DiffHead}_{\theta}(A_t^{(n)}, t; C^{(n-1)})$
    \State $\mathcal{L} \gets \mathcal{L} + \left\| \hat{A}_0^{(n)} - A_0^{(n)} \right\|_2^2$
\EndFor

\State \Return $\mathcal{L} / N$ \Comment{mask padded positions in practice}
\end{algorithmic}

\AlgoBottomRule
\end{minipage}
\hfill
\begin{minipage}[t]{0.485\textwidth}


\captionof{algorithm}{BitDance-LM sampling.}
\label{alg:bitdance_sample}

\AlgoTopRule

\small
\begin{algorithmic}[1]
\Require Prompt tokens $y_{\mathrm{prompt}}$, block size $m$, denoising schedule $1=t_K>\cdots>t_0=0$, maximum number of new blocks $N_{\mathrm{gen}}$
\Ensure Generated continuation $y_{\mathrm{gen}}$
\AlgoMidRule
\State Convert $y_{\mathrm{prompt}}$ into realized binary blocks $\{A_0^{(1)}, \ldots, A_0^{(n_0)}\}$ via $\phi$
\State $E^{(1:n_0)} \gets \operatorname{MLP}(A_0^{(1:n_0)})$
\State $C^{(n_0)}, \mathrm{KV} \gets \operatorname{BlockCausalLLM}(E^{(1:n_0)}; \mathcal{M}_m)$
\State $y_{\mathrm{gen}} \gets \emptyset$

\For{$j = 1, \ldots, N_{\mathrm{gen}}$}
    \State $n \gets n_0 + j$
    \State $A_{t_K}^{(n)} \sim \mathcal{N}(0, I_{m \times B})$

    \For{$k = K, K-1, \ldots, 1$}
        \State $\hat{A}_0^{(n)} \gets \operatorname{DiffHead}_{\theta}(A_{t_k}^{(n)}, t_k; C^{(n-1)})$
        \State $A_{t_{k-1}}^{(n)} \gets \dfrac{t_{k-1}}{t_k} A_{t_k}^{(n)}$
        \Statex \hspace{\algorithmicindent} $+ \left(1 - \dfrac{t_{k-1}}{t_k}\right) \hat{A}_0^{(n)}$
    \EndFor

    \State $\bar{A}_0^{(n)} \gets \operatorname{sign}(A_{t_0}^{(n)})$
    \State $y^{(n)} \gets \phi^{-1}(\bar{A}_0^{(n)})$
    \State Append $y^{(n)}$ to $y_{\mathrm{gen}}$ and truncate after EOS if needed

    \If{EOS appears in $y^{(n)}$}
        \State \textbf{break}
    \EndIf

    \State $E^{(n)} \gets \operatorname{MLP}(\bar{A}_0^{(n)})$
    \State $C^{(n)}, \mathrm{KV} \gets \operatorname{BlockCausalLLM}(E^{(n)}; \mathcal{M}_m, \mathrm{KV})$
\EndFor

\State \Return $y_{\mathrm{gen}}$
\end{algorithmic}

\AlgoBottomRule
\end{minipage}

\end{figure*}

\subsection{Inference}
\label{subsec:inference}

At inference time, decoding alternates between a causal backbone update and a blockwise denoising step, as summarized in Algorithm~\ref{alg:bitdance_sample}. 
Suppose blocks up to $n-1$ have already been realized. The backbone consumes the prompt and previously generated binary blocks, maintains a KV cache, and outputs the condition tensor $C^{(n-1)}$ for the next block. We then initialize the target block from Gaussian noise:
\begin{equation}
A^{(n)}_{t_K} \sim \mathcal{N}(\mathbf{0}, \mathbf{I}_{m \times B}),
\end{equation}
where $1=t_K > t_{K-1} > \cdots > t_0 = 0$ is a denoising schedule with $K$ steps. At each step, we predict the clean block and move the current state toward it:
\begin{align}
\hat{A}^{(n)}_0
&=
\operatorname{DiffHead}_\theta\!\left(A^{(n)}_{t_k},\, t_k;\, C^{(n-1)}\right), \\
A^{(n)}_{t_{k-1}}
&=
\frac{t_{k-1}}{t_k} A^{(n)}_{t_k}
+
\left(1-\frac{t_{k-1}}{t_k}\right)\hat{A}^{(n)}_0.
\label{eq:sampling_update}
\end{align}
After the final step, we project back to the binary hypercube:
\begin{equation}
\bar{A}^{(n)}_0 = \operatorname{sign}\!\left(A^{(n)}_{t_0}\right).
\label{eq:final_binary_block}
\end{equation}
The resulting binary codes are mapped back to token ids via $\phi^{-1}$ and appended to the prefix. The newly realized block is then fed through the same input MLP and backbone once to update the KV cache and produce the condition tensor for the following block.

The block size $m$ therefore acts as a direct parallelism knob. When $m=1$, BitLM reduces to standard autoregressive generation in binary space. For $m>1$, the model emits multiple future tokens per backbone update, while preserving causal dependence across blocks. In this sense, block-causal parallel generation is not a post-hoc decoding trick, but a native consequence of the model's binary denoising interface.
\begin{figure*}
    \centering
    \includegraphics[width=\linewidth]{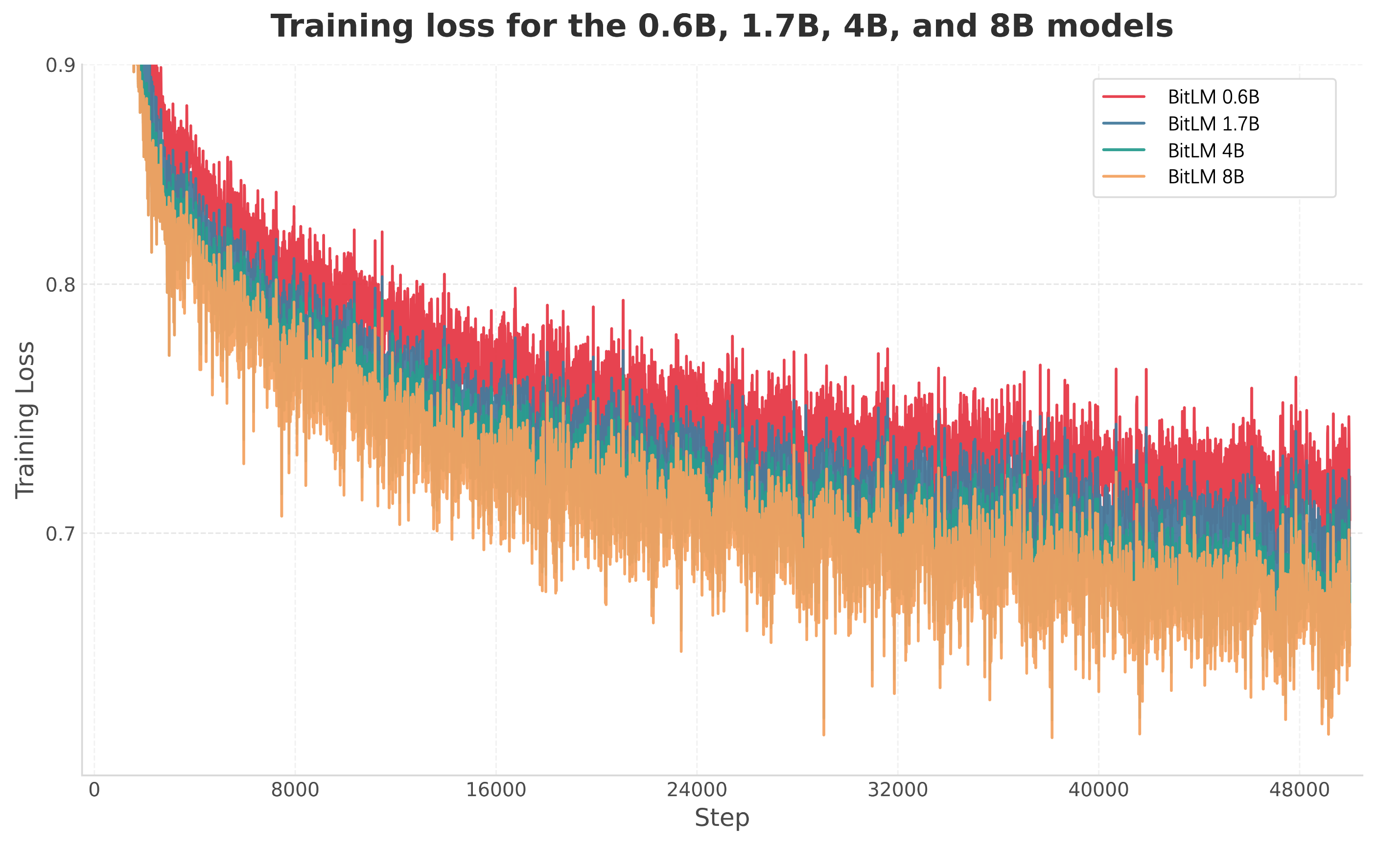}
    \caption{\textbf{Pretraining loss for the 0.6B, 1.7B, 4B, and 8B BitLM.}
    }
    \label{fig:size_abla}
\end{figure*}

\begin{figure*}
    \centering
    \includegraphics[width=\linewidth]{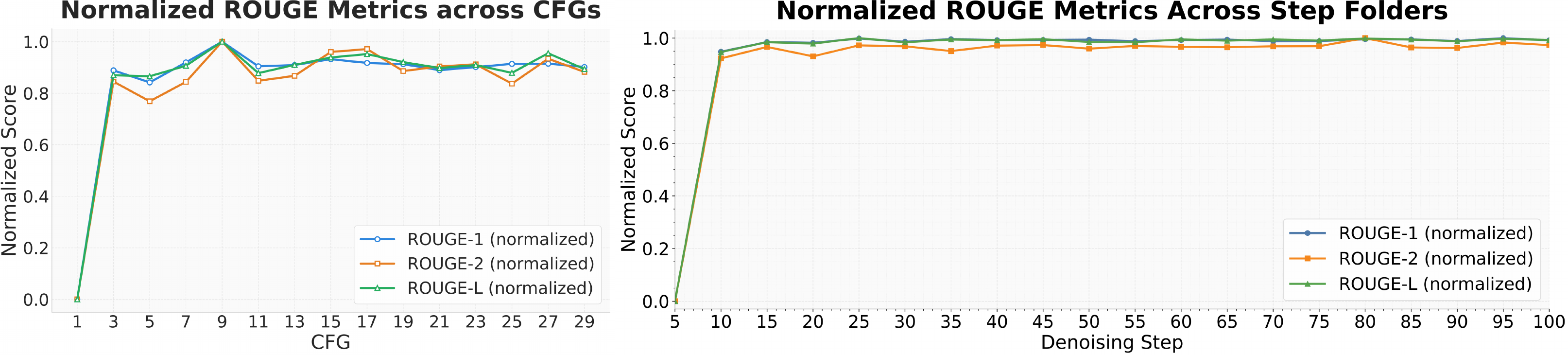}
    \caption{\textbf{Cfg and denoising step ablation of inference setting.}
    }
    \label{fig:cfg_step_abla}
\end{figure*}

\section{Experiments}
\label{sec:exp}

\textbf{Implementation.}
We pretrain BitLM on a subset of FineWeb's \citep{penedo2024fineweb} 350B tokens for 1 epoch and finetune it on the XSum \citep{narayan2018don} training set.
The LLM in BitLM follows the architecture of Qwen-3 \citep{yang2025qwen3}, 
while the diffusion head follows the architecture of BitDance's \citep{ai2026bitdance} diffusion head.
We set block size $m=4$.
During training, 
we employ the AdamW \citep{loshchilov2017decoupled} optimizer, setting $lr=1e-4$, $\beta_1=0.9$, $\beta_2=0.95$.
We concatenate $m$ \textit{$\langle bos\rangle$} tokens at the beginning of each sample and concatenate \textit{$\langle eos\rangle$} tokens at the end until the sample length is divisible by m.
For efficient training, 
we pack multiple samples into a single sequence, 
with each sequence containing 16384 tokens.
During inference,
we employ the ODE solver \citep{lipman2022flow} and set the denoising step $K=15$ and classifier-free guidance \citep{ho2022classifier} to $9.0$.

\begin{wraptable}{r}{0.46\linewidth}
    \vspace{-0.6em}
    \centering
    \small
    \setlength{\tabcolsep}{3.5pt}
    \renewcommand{\arraystretch}{1.08}
    \caption{\textbf{ROUGE results on XSum.}PT and FT refer to pretrained model and finetuned model respectively.}
    \label{tab:xsum_results}
    \resizebox{\linewidth}{!}{%
    \begin{tabular}{@{}lccc@{}}
        \toprule
        Method & R1 & R2 & RL \\
        \midrule
        Lead-3       & 16.30 & 1.60 & 11.95 \\
        PTGEN~\citep{see2017get}        & 29.70 & 9.21 & 23.24 \\
        PTGEN+COV~\citep{see2017get}    & 28.10 & 8.02 & 21.72 \\
        \midrule
        BitLM 8B \textit{w/.} LM Head (PT) & 10.06 & 2.64 & 8.78 \\
        BitLM 8B \textit{w/.} LM Head (FT) & 23.20 & 4.45 & 18.04 \\
        BitLM 8B \textit{w/.} Diff. Head (PT)   & 19.49 & 2.03 & 15.19 \\
        BitLM 8B \textit{w/.} Diff. Head (FT) & 26.05 & 6.44 & 20.12 \\
        \bottomrule
    \end{tabular}%
    }
    \vspace{-0.8em}
\end{wraptable}

\textbf{Scalability.}
As shown in Fig.~\ref{fig:size_abla},
We pretrain the 0.6B, 1.7B, 4B, and 8B versions of BitLM on FineWeb-350BT.
As the model size of BitLM increases, the training loss continues to decrease, demonstrating excellent scalability.
It is worth noting that we adopted the model architectures of Qwen-3 and BitDance, without incorporating any special designs.

\textbf{Denoising steps and classifier-free guidance.}
As shown in Fig.~\ref{fig:cfg_step_abla}LM
, 
when $K=15$ and the classifier-free guidance value is $9$, 
the fine-tuned BitLM 8B achieves the best performance on XSum.

\textbf{XSum summarization results.}
As shown in Tab.~\ref{tab:xsum_results}, it reports our results on XSum. After supervised fine-tuning, BitLM achieves 25.91/6.40/20.02 ROUGE-1/2/L. This result is above the Lead-3 baseline, showing that the model learns a non-trivial summarization mapping, but it remains below the pointer-generator baselines. We therefore view the current XSum result as promising but still preliminary rather than fully competitive.
At the same time, we view this experiment as informative rather than discouraging. The fact that BitLM can be pretrained at scale and then fine-tuned to obtain reasonable XSum performance suggests that replacing the vocabulary softmax with binary-space denoising is viable for downstream text generation. The remaining gap points to concrete directions for improvement, including stronger task-adaptive fine-tuning, better denoising schedules, adaptive block sizes, and hybrid designs that combine blockwise generation with summarization-specific alignment or copying mechanisms.

\section{Conclusion}

We presented BitLM, a language model that replaces the vocabulary softmax with conditional denoising in a fixed-length binary space. By coupling a causal LLM backbone with a diffusion head, BitLM jointly denoises multiple future tokens. This natively enables block-causal parallel generation without relying on post-hoc decoding tricks.

Empirically, BitLM scales smoothly during pretraining and adapts to downstream tasks, demonstrating that the large-vocabulary softmax is not the only viable output interface. While preliminary XSum results indicate that our binary formulation is not yet fully optimized for tasks requiring highly precise lexical realization, it opens a new design dimension: the geometry of the symbolic output space. We hope BitLM motivates future exploration into learned binary codes, adaptive blockwise schedules, and hybrid softmax--binary architectures that combine parallel generation with high-quality language modeling.
\section{Ethics Statement}

This work is intended to advance research on language modeling by exploring an alternative symbolic interface for text generation. Our experiments are conducted on publicly available datasets commonly used in prior research, including FineWeb for pretraining and XSum for summarization fine-tuning. We believe BitLM may support beneficial applications in language generation and model efficiency research, while we do not intend this framework for harmful or misleading use. As with other generative language models, downstream applications should be developed and deployed responsibly, with appropriate attention to fairness, safety, and legal compliance. The authors declare no conflict of interest.

\bibliography{main}
\bibliographystyle{colm2026_conference}

\appendix

\end{document}